\documentclass{article}
\usepackage{arxiv}

\usepackage[utf8]{inputenc} % allow utf-8 input
\usepackage[T1]{fontenc}    % use 8-bit T1 fonts
\usepackage{hyperref}       % hyperlinks
\usepackage{url}            % simple URL typesetting
\usepackage{booktabs}       % professional-quality tables
\usepackage{amsmath, amsfonts}       % blackboard math symbols
\usepackage{nicefrac}       % compact symbols for 1/2, etc.
\usepackage{microtype}      % microtypography
\usepackage{graphicx}
\usepackage{cite}
\usepackage{doi}
\usepackage{float}
\usepackage{subcaption}

\title{Survey and Evaluation of Converging Architecture \\ in LLMs based on Footsteps of Operations}

\author{
    \href{https://orcid.org/0009-0008-9306-9301}{{\bf Seongho Kim} \textsuperscript{\includegraphics[scale=0.05]{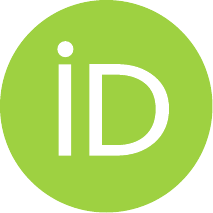}1}}, 
    \href{https://orcid.org/0009-0006-5280-3392}{{\bf Jihyun Moon} \textsuperscript{\includegraphics[scale=0.05]{orcid.pdf}1}}, 
    \href{https://orcid.org/0009-0002-6909-8942}{{\bf Juntaek Oh} \textsuperscript{\includegraphics[scale=0.05]{orcid.pdf}1}}, 
    \href{https://orcid.org/0009-0009-2016-6714}{{\bf Insu Choi} \textsuperscript{\includegraphics[scale=0.05]{orcid.pdf}1}}, 
    \href{https://orcid.org/0000-0002-1502-5353}{{\bf Joon-Sung Yang} \textsuperscript{\includegraphics[scale=0.05]{orcid.pdf}1, 2, 3}} \\
    {\small
    \textsuperscript{1}Department of Electrical and Electronic Engineering,
    \textsuperscript{2}Department of Systems Semiconductor Engineering,} \\
    {\small
    \textsuperscript{3}BK21 Graduate Program in Intelligent Semiconductor Technology, Yonsei University, Seoul, South Korea}
}

\renewcommand{\headeright}{}
\renewcommand{\undertitle}{}
\renewcommand{\shorttitle}{}

\begin{document}

\maketitle

\begin{abstract}
Large Language Models (LLMs), emerging from advancements in Natural Language Processing (NLP) tasks, allow chatbots to provide more sophisticated and human-like text generation by leveraging their vast model sizes, often exceeding billions of parameters, to provide deep knowledge across various domains.
Hence, they are becoming more integrated into our daily lives, serving as personal assistants or even as experts across various domains.
Initial language models relied on rule-based systems and early neural networks like Recurrent Neural Networks (RNNs).
However, it remained an issue to handle long-term dependency and to understand context over extended conversation.

The advent of the Attention mechanism and Transformer architecture enables contextually natural text generation and compresses the burden of processing entire source information into singular vectors.
Based on these two main ideas, model sizes gradually increases to accommodate more precise and comprehensive information, leading to the current state-of-the-art LLMs being very large, with parameters around 70 billion.
As the model sizes are growing, the demand for substantial storage and computational capacity increases.
This leads to the development of high-bandwidth memory and accelerators, as well as a variety of model architectures designed to meet these requirements.
We note that LLM architectures have increasingly converged. 
This paper analyzes how these converged architectures perform in terms of layer configurations, operational mechanisms, and model sizes, considering various hyperparameter settings.

In this paper, we conduct a concise survey of the history of LLMs by tracing the evolution of their operational improvements.
Furthermore, we summarize the performance trends of LLMs under various hyperparameter settings using the RTX 6000, which features the state-of-the-art Ada Lovelace architecture.
We conclude that even the same model can exhibit different behaviors depending on the hyperparameters or whether it is deployed in server or edge environments.
\end{abstract}

% keywords can be removed

\section{Introduction}
In the past few years, since the release of the transformer model \cite{vaswani2017attention}, AI models have evolved rapidly.
Among these models, LLMs have garnered significant interest and made substantial progress, particularly due to the popularity of ChatGPT.
There are a variety of model options available on the market, with open-source LLMs being particularly popular due to their cost-effectiveness and extensive customizability.
One notable advantage of open-source LLMs is their cost-efficiency, offering free access to high-performance models.
Furthermore, these models provide the flexibility of customization and optimization, allowing users to adapt the models to their particular requirements.
Notably, Meta's Llama and Google's Gemma have attracted considerable interest.

Llama \cite{touvron2023llama1} was unveiled with the initial model available in four versions, ranging from 6.7B to 65.2B parameters.
Building on this, Llama 2 \cite{touvron2023llama2} was released in three versions: 7B, 13B, and 70B, with a 40\% increase in the size of its pre-training data.
The sequence length was doubled over the previous model, and the 70B version was the first to incorporate Grouped Query Attention (GQA), demonstrating much stronger performance.
Additionally, Meta released Code Llama based on Llama 2 and announced Llama 3.

Meanwhile, Google DeepMind introduced their open-source LLM, Gemma \cite{team2024gemma}.
This lightweight model was released in two sizes with 7B and 2B parameters, targeting on-device applications such as desktops or mobile devices.
Following this, Google DeepMind updated Gemma to version 1.1, maintaining the model size while enhancing performance.
They have continuously developed and updated the Gemma family, including the release of RecurrentGemma and CodeGemma.

Efforts to accelerate these models have also been significant.
One notable development is the release of TensorRT-LLM, an open-source library that accelerates the inference performance on NVIDIA GPUs.
As LLMs have evolved and grown in size, their computational cost has increased significantly, making them challenging to operate without modern techniques.
To address this, TensorRT-LLM \cite{tensorrt_llm} provides a comprehensive library that can be compiled and optimized for inference.
The library includes a number of optimization techniques, such as kernel fusion and quantization, as well as runtime optimizations such as C++ implementations, KV caching, continuous flight batching, paged Attentions, and an intuitive Python API for defining and building new models.
It also supports multi-GPU and multi-node inference, and is constantly being updated for the latest models.
Furthermore, it is compatible with the latest NVIDIA GPU architecture.

However, since the announcement of Llama, the structural development of these models has gradually converged, and the differences between models often depend on the dataset and the specific operations used.
Significant changes have occurred in computational methods, Attention types, and activation functions.
%Specifically, after the use of Multi-Head Attention (MHA) in Transformers, variations such as Grouped Query Attention (GQA) and Multi-Query Attention (MQA) have been employed.
%Additionally, activation functions have evolved from Rectified Linear Unit (ReLU) to more advanced functions like Gaussian Error Linear Unit (GeLU) and Swish.
As a result, LLMs have been differentiated into server AI, edge AI, and on-device AI, depending on the purpose of the model and the resources that can be allocated.
Therefore, we analyze the latest open-source LLMs to understand their current technical limitations and provide directions for developing better performance and efficiency models.
Among the three focuses, model performance, learning time, and inference time, we selected the inference time as the primary target.

In this paper, we survey the evolution of language models from Recurrent Neural Networks (RNNs) to LLMs, focusing on their key operations and architectures.
In addition, we analyze the architecture of modern open-source LLMs such as Gemma and Llama.
Specifically, we examine the inference process of these models on high-performance GPUs, analyze the weight of each kernel, and identify the bottlenecks in current hardware.

\begin{figure*}[t]
\centering
\includegraphics[width=0.85\linewidth]{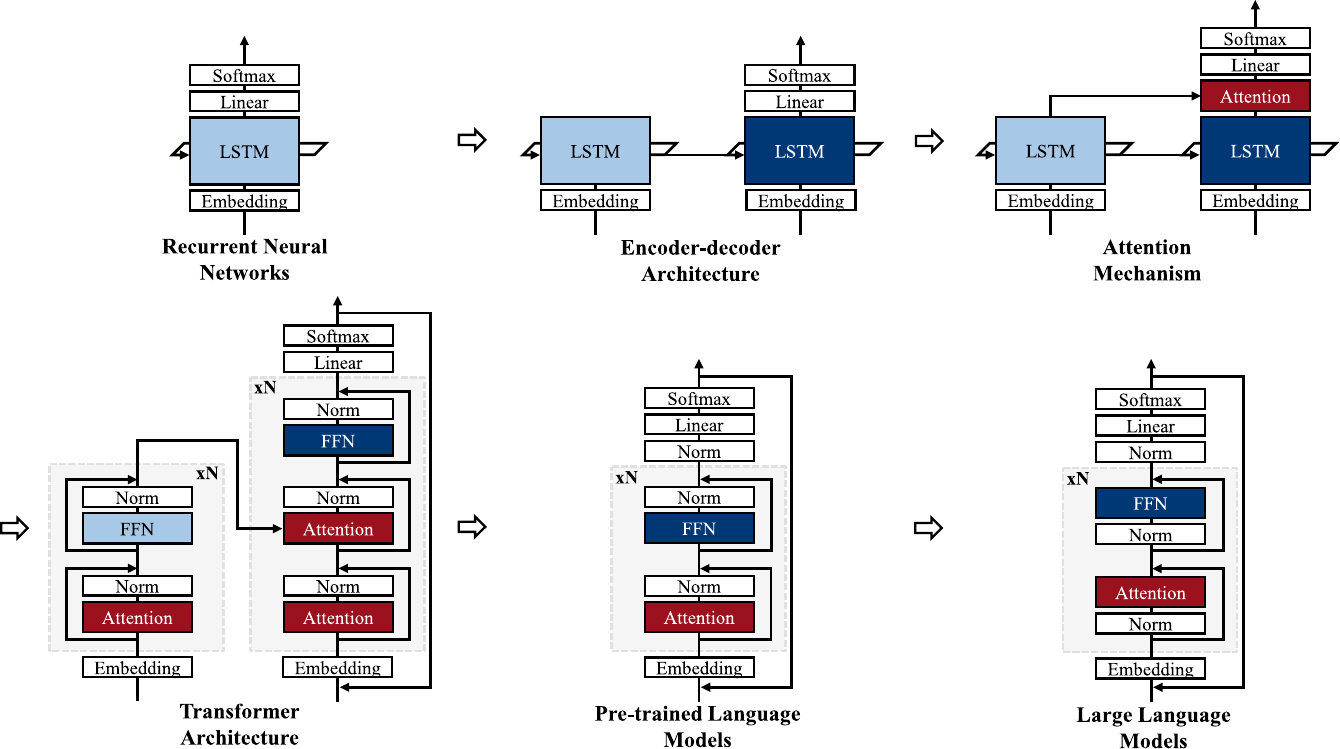}
%\captionsetup{font=small}
\caption{Architecture overview from Recurrent Neural Networks (RNNs) to Large Language Models (LLMs)}
\label{fig.architecture_overview}
\end{figure*}

\section{Development of Architecture}\label{Footsteps for LLM}
In this section, we provide a overview of the development process of previous language models targeting NLP tasks, with a particular focus on architecture (Figure \ref{fig.architecture_overview}). The major milestones from RNNs to modern LLMs are summarized.

\subsection{Recurrent Neural Networks}
In earlier stages, linguistic processing techniques were limited to rule-based or statistical approaches due to issues such as complexity or language specificity. 
Past neural networks had a structure in which only perceptrons were simply placed, and therefore the context of language could not be considered. 
However, with the advent of RNNs that are capable of identifying temporal correlations by their circular structure, neural networks have begun to be actively utilized in the field of language. 
In particular, Long Short-Term Memory (LSTM) \cite{hochreiter1997long}, which has improved the structure of existing RNNs that are prone to gradient vanishing problems in the long term, has been extensively adopted in various early language models.

\subsection{Encoder-decoder Architecture}
\cite{sutskever2014sequence} proposes an encoder-decoder architecture composed of two LSTMs to address the limitations of the existing model, which could only accommodate the fixed-length input. 
The encoder LSTM sequentially encodes the input to create a long fixed-dimensional vector, and the decoder LSTM decodes this to produce the output. 
This structure responds to varying-length input by receiving input tokens sequentially, and also improves performance in NLP tasks due to the long memory of LSTM.

\subsection{Attention Mechanism}
The encoder-decoder structure converts the entire input sentence into a fixed-length vector, thus creating the bottleneck issue. 
Performance degradation occurs as the input sentences become longer. 
In order to address this issue, \cite{bahdanau2014neural} proposes the Attention technique that focuses on the highly relevant part of the context word generated by the encoder for each output from the decoder.
This alleviates the burden that arises when the model compresses the entire source information into a singular vector and enhances comprehension of the contextual nuances.

\subsection{Transformer Architecture}
RNNs are inherently sequential, precluding parallelized computation within training datasets.
They also generate the memory burden as the sequence becomes longer, limiting batching across examples. 
The transformer structure proposed in \cite{vaswani2017attention} maintains the encoder-decoder structure and replaces all RNNs with attention mechanism and fully connected feed-forward network. 
By removing recurrence, the model facilitates much higher levels of parallelization, improving speed and performance simultaneously in training and inference.
The transformer has become the base model for modern language models.

\subsection{Pre-trained Language Models}
GPT-1 \cite{radford2018improving} and BERT \cite{devlin2018bert} define the model structure with only a decoder or an encoder, respectively, and further increase internal parameters such as the number of layers and dimension size. 
They also strengthen pre-training and fine-tuning techniques, such as integrating unsupervised pre-training and supervised fine-tuning, or incorporating the "Masked Language Model" and "Next Sentence Prediction" tasks.
Therefore, the model's context understanding ability was greatly improved, achieving the State-of-the-Art (SOTA) on benchmarks such as General Language Understanding Evaluation (GLUE).

\subsection{Large Language Models}
Facebook AI team discovers that BERT is significantly under-trained by the previous methods, and reveals RoBERTa \cite{liu2019roberta} with improved performance. 
The team upgrades the model with several modifications, such as removing previous prediction tasks and expanding input sequence, batch sizes, training dataset and duration.
Likewise, Google AI Language team reveals GPT-3 \cite{brown2020language}, by removing fine-tuning, expanding parameter to 175 billion and adopting few-shot learning on a vast amount of datasets. 
Therefore, GPT-3 could achieve strong performance on various NLP tasks.
As a result of these previous studies, various language models improve performance by increasing parameters and datasets, leading to the development of modern LLMs. 
Recent high-performance LLMs follow a transformer-based decoder-only architecture.

\begin{figure}[t]
\centering
\includegraphics[width=0.85\columnwidth]{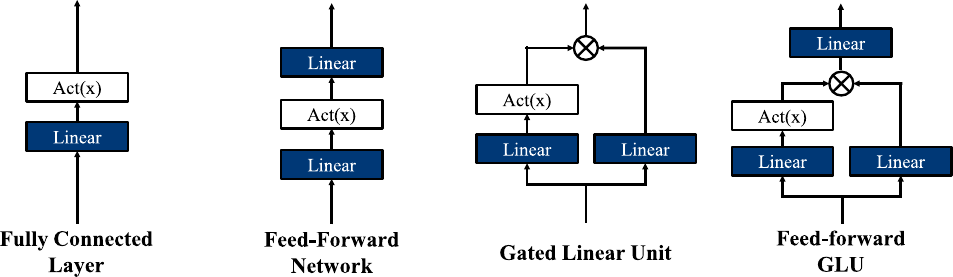}
\caption{Fully Connected layer (FC layer), Feed-forward Neural Network (FFN), Gated Linear Unit (GLU) and Feed-forward GLU.}
\label{fig.neural_network}
\end{figure}

\section{Development of Operation}
This section explains the primary operations performed in each layer of modern models.
It also provides definitions of the techniques used in each layer and describes their development processes.

\subsection{Feed-forward Layer}
All neural networks begin with a component known as a perceptron. 
The perceptron generates an output by adding bias to the cumulative sum of multiple inputs multiplied by each weight and then passing it through the activation function. 
A Fully Connected layer (FC layer) is a form in which the perceptrons are aligned in a line. 
Upon analysis in a vertical direction, it can be observed that it consists of a linear transformation and an activation function. 
The transformer model defines a Feed-Forward Network (FFN) in which an activation function exists between two linear transformations. 
Gated Linear Unit (GLU) \cite{dauphin2017language} is proposed to resolve the gradient vanishing problem and improve prediction accuracy. 
GLU is a structure that converts the input into two linear transformations and performs a component-wise product, one of which is passed through an activation function. 
Furthermore, \cite{shazeer2020glu} introduces GLU variants that apply several modern activation functions to GLU, as well as defining FFN variants (Feed-forward GLU) that apply GLU to the existing transformer FFN. 
The operations are described in Figure \ref{fig.neural_network}, and also expressed as:

\begin{equation}
\operatorname{FC}(x, W, b)=\operatorname{Act}(xW+b)
\end{equation}
\begin{equation}
\operatorname{FFN}\left(x, W, W_2, b, b_2\right)=\operatorname{Act}(xW+b)W_2+b_2
\end{equation}
\begin{equation}
\operatorname{GLU}(x, W, V, b, c)=\operatorname{Act}(xW+b) \otimes(xV+c)
\end{equation}
\begin{equation}
\begin{split}
\operatorname{FFN_{GLU}}\left(x, W, V, W_2\right) \\
=(\operatorname{Act}(xW) \otimes xV)W_2
\end{split}
\end{equation}

where $x$ is the input vector, $W$, $V$, $W_2$ are weight matrices, and $b$, $c$, $b_2$ are bias vectors. 
\(\operatorname{Act}(x)\) is a generalized form of the activation function. 
Note that the bias terms are omitted in feed-forward GLU.

\subsection{Attention Layer}
The attention mechanism is a method that focuses on and refers to highly relevant parts of the entire input sentence of the encoder during each decoder output process.
It computes the similarity between the input Query (Q) and all Keys (K), then applies this similarity to the corresponding Value (V) for each key, and cumulatively adds these results to return the final outcome.
Scaled Dot-Product Attention (SDPA), a representative form among variations of Attention, goes through the following process: after dot-producing the Q and K, SPDA calculates the "Attention score" by multiplying it by the scaling factor, \(1/\sqrt{d_k}\). 
Next, SDPA takes the softmax, calculates the "Attention distribution", and products it with V to obtain the "Attention value".

\begin{figure}[t]
\centering
\includegraphics[width=0.85\columnwidth]{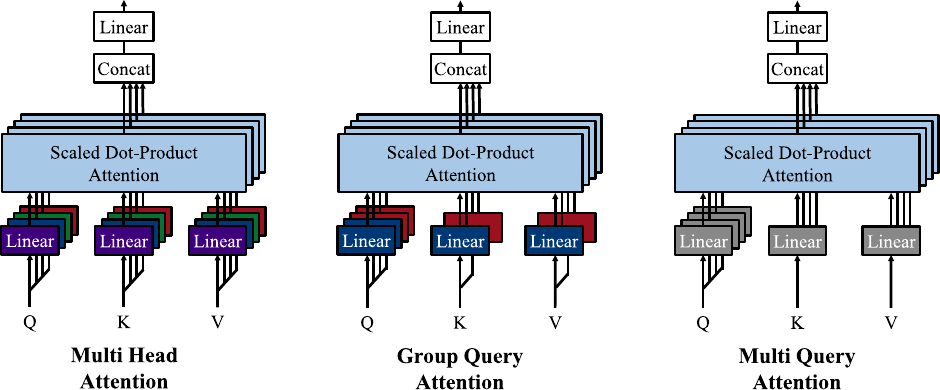}
\caption{Multi Head Attention (MHA), Group Query Attention (GQA) and Multi Query Attention (MQA).}
\label{fig.mha_gqa_mqa}
\end{figure}
 
When performing a linear projection on a multi-head rather than a single Attention, the overall performance is improved since each head is able to comprehend the context in subspaces of different positions and complement each other. 
This method is referred to as Multi-head Attention (MHA), and the transformer model employs SDPA and MHA simultaneously. 
Although \cite{shazeer2019fast} acknowledges the performance of MHA, it proposes Multi-Query Attention (MQA) by focusing on the memory burden that arises when repeatedly loading K and V, which are large tensors. 
MQA can significantly reduce memory bandwidth by sharing a single KV head with all Q heads, however, it has the disadvantage of deteriorating model performance. 
Based on this, \cite{ainslie2023gqa} proposes Grouped Query Attention (GQA), which comprises MHA and MQA by allocating a single KV head at each Q head group.

\newpage

Three different attention techniques are shown in Figure \ref{fig.mha_gqa_mqa}, and can also be described as:

\begin{equation}
\operatorname{SDPA}(Q, K, V)=\operatorname{softmax}\left(\frac{Q K^T}{\sqrt{d_k}}\right) V
\end{equation}
\begin{equation}
\begin{split}
\operatorname{MHA}(Q, K, V)=\text {Concat}\left(\text {head}_1, \ldots, \text {head}_{\mathrm{h}}\right) W^O \\
%\end{equation}
%\begin{equation}
\text {where head}_{\mathrm{i}}=\operatorname{SDPA}\left(Q W_i^Q, K W_i^K, V W_i^V\right)
\end{split}
\end{equation}
\begin{equation}
\text {if GQA, head}_{\mathrm{i}}=\operatorname{SDPA}\left(Q W_i^Q, K W_j^K, V W_j^V\right)
\end{equation}
\begin{equation}
\text {if MQA, head}_{\mathrm{i}}=\operatorname{SDPA}\left(Q W_i^Q, K W^K, V W^V\right)
\end{equation}

where $Q$, $K$ and $V$ are a query, key, and value matrix, respectively. 
The dimension of the query and the key is both $d_k$, and the dimension of the value is $d_v$. 
In the equation, parameter matrices are defined as \(W_i^Q \in \mathbb{R}^{d_{\mathrm{model}} \times d_k}\), \(W_i^K, W_j^K, W^K \in \mathbb{R}^{d_{\mathrm{model}} \times d_k}\), \(W_i^V, W_j^V, W^V \in \mathbb{R}^{d_{\mathrm{model}} \times d_v}\), and \(W^O \in \mathbb{R}^{hd_v \times d_{\mathrm{model}}}\).

Flash Attention \cite{dao2022flashattention} is a mechanism that improves the memory access of the traditional attention mechanism to increase computational efficiency. 
The memory complexity of self-attention operation increases quadratically with the increase in sequence length, which is time consuming and requires a lot of data movement between memories. 
To address the challenge, the flash attention uses an IO-aware algorithm that optimizes memory access considering the hierarchy of GPU memory. 
It divides queries, keys, and values into blocks and places them in SRAM for processing, stores intermediate computation results in SRAM for recomputation, and utilizes a tiling technique to minimize access to the GPU HBM. 
This technique optimizes memory access and allows us to efficiently handle long sequences. 
Since then, Flash Attention v2 has been released with several improvements. 
First, the number of GEMV operations was reduced to enhance computation efficiency, and single-head computations were parallelized to increase GPU occupancy. 
Furthermore, the work was distributed between warps within thread blocks to reduce access to shared memory. 
These methods optimized the computation of attention to a level of efficiency comparable to that of GEMM operation.

\subsection{Normalization Layer}
During training, the phenomenon called the "internal covariate shift" occurs whenever the parameter is converted, leading to a decrease in learning performance. 
Normalization techniques are proposed to address this issue. 
Batch Normalization (BatchNorm) \cite{ioffe2015batch} normalizes the input by calculating the average and variance of multiple mini batches. 
Although batch normalization is effective for CNNs, it is not suitable for language models where the length of the sequence varies. 
Therefore, Layer Normalization (LayerNorm) \cite{ba2016layer}, which normalizes within a single batch, is presented. 
Afterwards, \cite{zhang2019root} claims that the process of re-centering and re-scaling of layer normalization is unnecessary.
It proposes Root Mean Square Normalization (RMSNorm) which normalizes based on RMS without re-centering and re-scaling of layer normalization.
%Modern language models actively adopt the advantages of RMS normalization.
The three different normalization techniques introduced are expressed as:

\begin{equation}
\hat{x}_{\text {BatchNorm}}=\frac{x-E_{\text {mini-batch}}(x)}{\sqrt{\operatorname{Var}_{\text {mini-batch}}(x)+\epsilon}} \cdot \gamma+\beta
\end{equation}
\begin{equation}
\hat{x}_{\text {LayerNorm}}=\frac{x-E_{\text {features}}(x)}{\sqrt{\operatorname{Var}_{\text {features}}(x)+\epsilon}} \cdot \gamma+\beta
\end{equation}
\begin{equation}
\hat{x}_{\text {RMSNorm}}=\frac{x}{\sqrt{E_{\text {features}}\left(x^2\right)+\epsilon}} \cdot \gamma
\end{equation}
where \(\operatorname{E}_{\text {mini-batch}}(x)\) and \(\operatorname{Var}_{\text {mini-batch}}(x)\) are the mean and variance, computed per feature over the mini-batch, and \(\epsilon\) is the constant for numerical stability. 
\(\gamma\) and \(\beta\) are scaling and shifting learnable parameters, respectively. \(\operatorname{E}_{\text {features}}(x)\) and \(\operatorname{Var}_{\text {features}}(x)\) are the mean and variance, computed over the feature dimension.

\begin{figure}[t]
    \centering
    \begin{subfigure}{0.48\columnwidth}
        \centering
        \includegraphics[width=1\columnwidth]{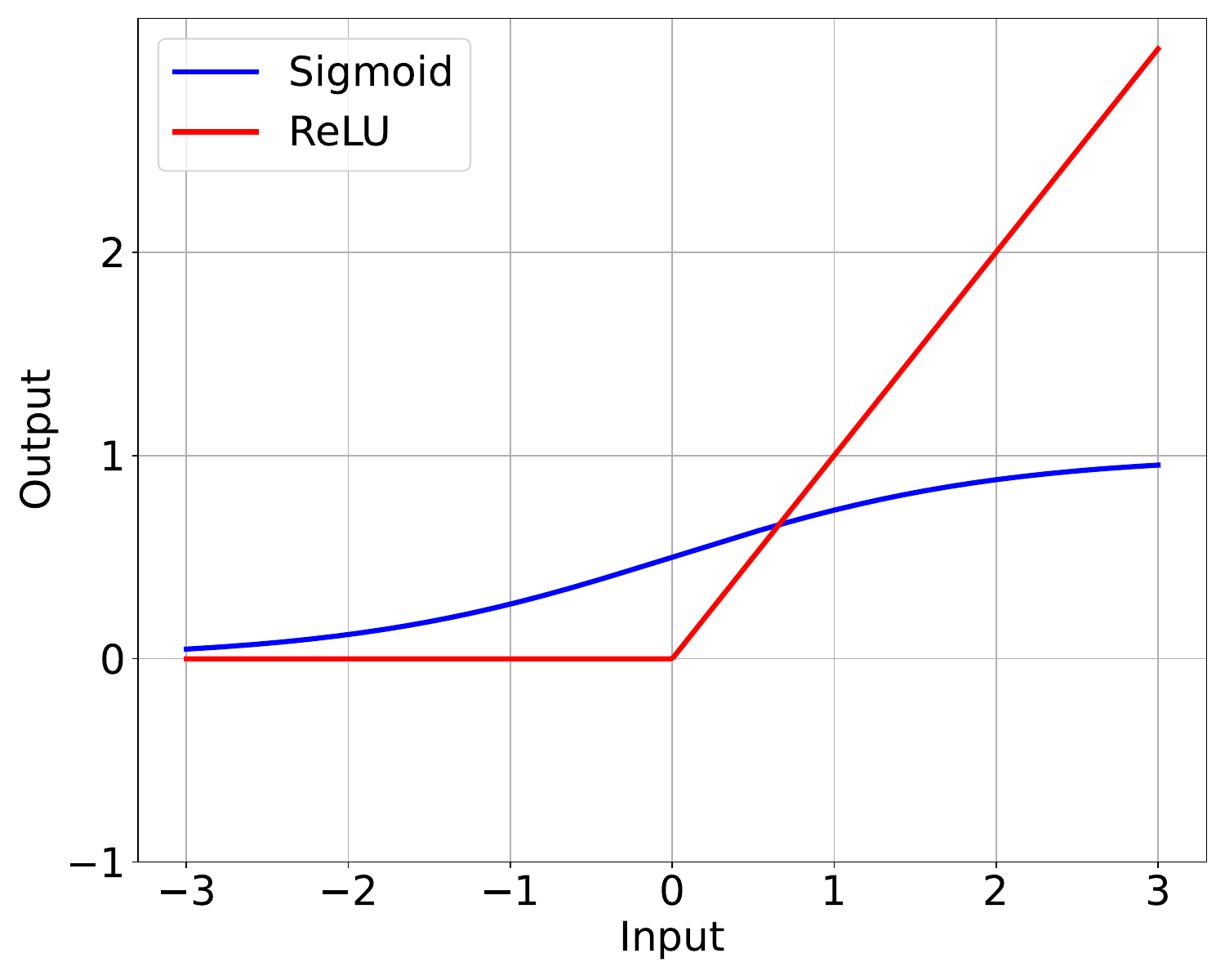}
        %\captionsetup{font=small}
        \caption{Sigmoid and ReLU activation function}
        \label{fig.activation_function1}
    \end{subfigure}
    \hfill
    \begin{subfigure}{0.48\columnwidth}
        \centering
        \includegraphics[width=1\columnwidth]{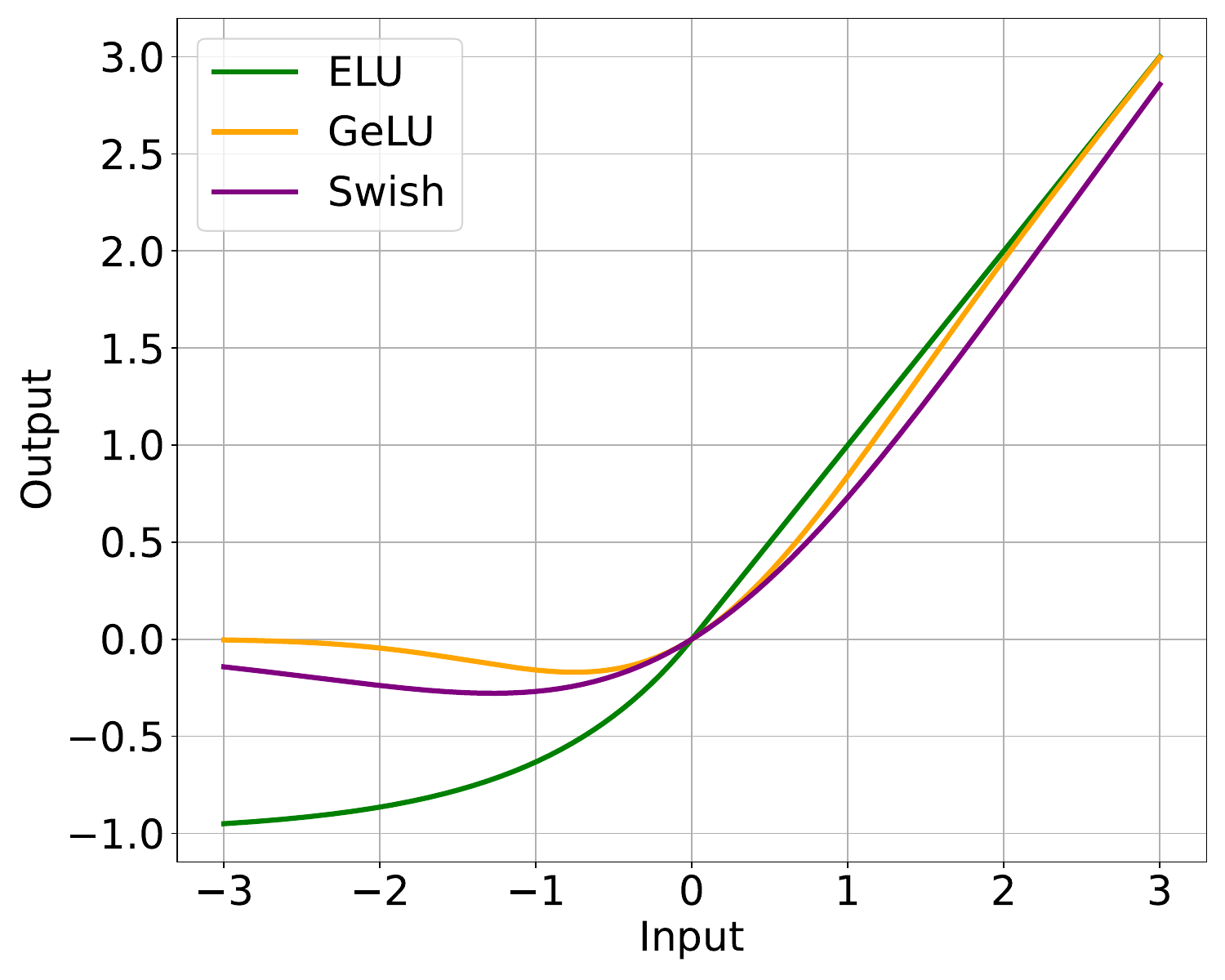}
        %\captionsetup{font=small}
        \caption{ReLU, ELU, GeLU and Swish activation function}
        \label{fig.activation_function2}
    \end{subfigure}
    \caption{Activation functions}
    \label{fig.combined activation functions}
\end{figure}

\subsection{Activation Function}
The activation function has also been developed to enhance performance along with the structure. 
Sigmoid is an activation function that is commonly utilized in probability representation tasks. 
It exhibits a distinctive profile, comprising a high slope in the center and a gradual decline toward both sides. 
However, a gradient vanishing problem emerges during back propagation, in which learning performance deteriorates as the distance from the output layer increases. 
In contrast, Rectified Linear Unit (ReLU) \cite{nair2010rectified} has a formula that rectifies the negative input of a linear function. 
In the case of positive input during back propagation, the gradient vanishing problem is eliminated, yet the update is not performed for negative input, resulting in the "dying ReLU" phenomenon in which most parameters become 0. 
Consequently, several activation functions that give a small gradient to the negative input have been proposed, including Exponential Linear Unit (ELU) \cite{clevert2015fast}, Gaussian error Linear Unit (GeLU) \cite{hendrycks2016gaussian}, and Swish \cite{ramachandran2017swish}.
The five activation functions above are plotted in Figure \ref{fig.combined activation functions}.
They sacrifice the computational efficiency of ReLU, and take improved performance. 
Several activation functions are expressed as:

\begin{equation}
\operatorname{sigmoid}(x)=\frac{1}{1 + e^{-x}}
\end{equation}
\begin{equation}
\operatorname{ReLU}(x)=\max (0, x)
\end{equation}
\begin{equation}
\operatorname{ELU}_{\alpha=1}(x)=
\begin{cases}
x & \text{for } x=0 \\
e^x-1 & \text{for } x<0
\end{cases}
\end{equation}
\begin{equation}
\operatorname{GeLU}(x)=x \cdot \Phi(x)=x \cdot \frac{1}{2}[1+\operatorname{erf}(x / \sqrt{2})]
\end{equation}
\begin{equation}
\operatorname{Swish}_{\beta=1}(x)=\frac{x}{1 + e^{-x}}
\end{equation}
where both ELU and Swish are described under the assumption of $\alpha=1$ and $\beta=1$. 
In addition, \(\operatorname{erf}(x)\) inside GeLU is defined as \(\operatorname{erf}(x)=\frac{2}{\sqrt{\pi}} \int_0^x e^{-t^2} \mathrm{~d} t\). 

\begin{equation}
\operatorname{softmax}\left(x_i\right)=\frac{e^{x_i}}{\sum_{j=1}^n e^{x_i}}
\end{equation}

The softmax function transforms input values into a set of output values, allowing classification of the input into one of multiple classes. 
The properties of applying the exponential function to inputs and normalizing them enable a probabilistic interpretation.
Therefore, the softmax function is generally placed in the last layer of models and plays a role in determining the output token.

\subsection{Residual Connection}
When training a neural network, it is commonly expected that models with a higher number of layers, the so-called "deep" models, would yield better performance. 
However, there is a phenomenon where the accuracy decreases as the depth of the model increases. 
To address this challenge, the concept of a "deep residual learning framework" is introduced in \cite{he2016deep}. 
This framework significantly improves the degradation problem by introducing "short connections" or "residual connections," which skip one or more layers in the existing structure. 
This solution has become a crucial component in various neural network architectures, including CNNs and language models, and remains essential to date. 

\subsection{Embedding and Encoding}
Embedding refers to the process of converting input text into a vector. 
Great deal of research has been conducted since model performance is significantly influenced by the manner in which words are expressed. 
As the dataset becomes vast, the one-hot encoding method is abandoned and the learned embedding method, in which semantic relationships are learned through language models, is primarily employed. 
From the transformer model, the concept of positional embedding is initiated by including positional encoding in embedding. 
This is because when replacing RNNs, the positional information of the token within the sequence must be included. 
Sinusoidal position encoding, used here, is a method of encoding absolute positions expressed as trigonometric functions of different frequencies into an embedding vector. 
\cite{su2024roformer} presents Rotary Position Embedding (RoPE), which encodes absolute position in embedding with a rotation matrix and adds relative position dependency in the self-Attention equation. 
This method is widely known to be efficient because it only performs a series of rotation operations. 
Furthermore, the relative distance and angle between tokens are preserved, allowing the model to easily learn relative positional information.

\begin{table}[b]
\small
\caption{Parameters of Llama and Gemma}
\centering
\renewcommand{\arraystretch}{1.2}
\begin{tabular}{lcccc}
\toprule
& \multicolumn{2}{c}{\textbf{Llama Parameters}} & \multicolumn{2}{c}{\textbf{Gemma Parameters}} \\
\cmidrule(lr){2-3} \cmidrule(lr){4-5}
& \textbf{2-7B (MHA)} & \textbf{3-8B (GQA)} & \textbf{1-2B (MQA)} & \textbf{1-7B (MHA)} \\
\midrule
Embedding size & 4096 & 8192 & 2048 & 3072 \\
Vocabulary size & 32000 & 32000 & 256128 & 256128 \\
Number of layers & 32 & 80 & 18 & 28 \\
Feed-forward dimensions & 11008 & 28672 & 32768 & 49152 \\
Number of heads & 32 & 64 & 8 & 16 \\
Number of KV heads & 32 & 8 & 1 & 16 \\
\bottomrule
\end{tabular}
\label{tab:llama_gemma}
\end{table}

\section{Architecture of Modern LLMs}
In this section, we provide a snapshot of modern open-source LLMs by analyzing the architecture (Section \ref{Architecture}) and layers (Section \ref{Layer Analysis}) of Llama and Gemma. 
We use Llama2-7B, Llama3-8B, Gemma-2B, and Gemma-7B models (Gemma 1.0 version), therefore, we could analyze MHA, GQA, and MQA in the same environment. Table \ref{tab:llama_gemma} provides a summary of the parameters for Llama and Gemma utilized in the experiments conducted in this paper.

\newpage

\begin{figure}[t]
\centering
\includegraphics[width=0.8\columnwidth]{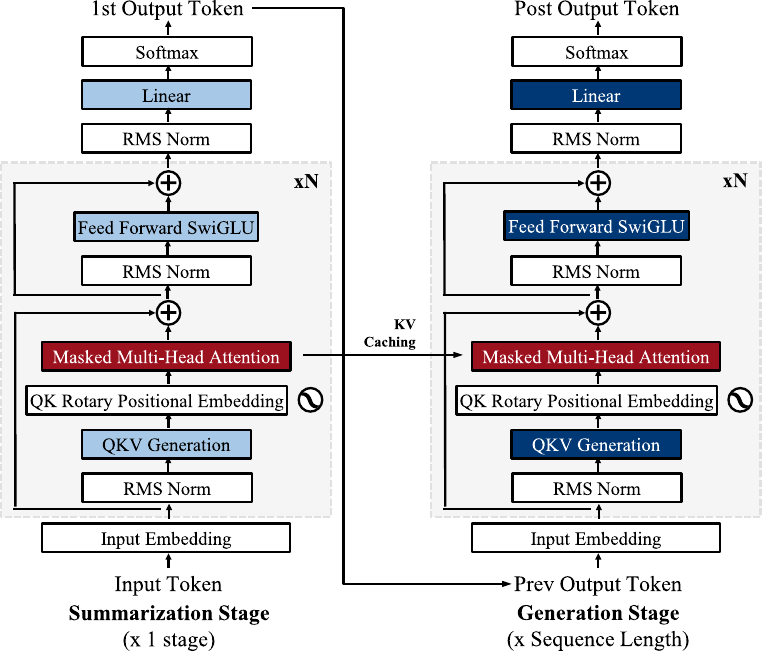}
\caption{Architecture of Llama (when GLU is SwiGLU) and Gemma (when GLU is GeGLU).}
\label{fig.archi}
\end{figure}

\subsection{Architecture Analysis}\label{Architecture}
As shown in Figure \ref{fig.archi}, the model Llama architecture follows decoder-only transformer-based architecture. 
It applies pre-normalization of RMSNorm, feed-forward SwiGLU, and RoPE. 
GQA is applied in the Llama 3-8B model. 
The architecture of Gemma model is mostly consistent with Llama, except that GeGLU is used instead of SwiGLU for activation function in feed-forward layer. 
MQA is also applied in the case of the Gemma-2B model.
In Figure \ref{fig.archi}, the colors are applied with respect to the types of kernels generated when each layer is executed, as observed in Figure \ref{fig:batch sweep} of the subsequent section.
The layers colored in light blue (summarization stage) and dark blue (generation stage) represent Matrix Multiplication layers, with the GEMM or GEMV operations. 
The attention layer, which is critical for the advancement and differentiation of models, is marked in red.

\subsection{Layer Analysis}\label{Layer Analysis}
Through the analysis of open-source models, we can summarize the variations in the torch size of each layer.
The torch size of the summarization, the generation stage, and the symbols can be organized as Table \ref{tab:sumgen layer}.
After the embedding process, computations are performed using the predetermined dimension size. 
In the attention block, the torch is split according to the number of heads appropriate for the structure, and it is recombined to its original configuration once the Attention block is complete.
In the generation stage, the KV caching size, C, is updated to L after the attention block. 
Therefore, as multiple attention blocks are processed, the size of C continuously increases in each output execution.

\begin{table*}[t]
\scriptsize
\caption{Tensor Dimension of Summarization Stage and Generation Stage}
\raggedleft
\renewcommand{\arraystretch}{1.2}
\begin{tabular}{llll}
\toprule
\textbf{Summarization Stage Layer} & \textbf{Dimension Calculation} & \textbf{Generation Stage Layer} & \textbf{Dimension Calculation} \\
\midrule
Token embedding & $(B, S, V) \times (V, E) = (B, S, E)$ & Token embedding & $(B, 1, V) \times (V, E) = (B, 1, E)$ \\
Pre-normalization (RMSNorm) & Torch dimension is not changed & Pre-normalization (RMSNorm) & Torch dimension is not changed\\
Query projection (weight GEMM) & $(B, S, E) \times (E, E) =$ & Query projection (weight GEMM) & $(B, 1, E) \times (E, E) =$ \\ & \hfill $(B, S, E) \rightarrow (B, H, S, A)$ & & \hfill $(B, 1, E) \rightarrow (B, H, 1, A)$\\
Key projection (weight GEMM) & $(B, S, E) \times (E, E) = (B, S, E)$ & Key projection (weight GEMM) & $(B, 1, E) \times (E, E) = (B, 1, E)$ \\ & \hfill $\rightarrow (B, h \rightarrow H, S, A)$ & & \hfill $\rightarrow (B, h \rightarrow H, L, A)$ \\
Value projection (weight GEMM) & $(B, S, E) \times (E, E) = (B, S, E)$ & Value projection (weight GEMM) & $(B, 1, E) \times (E, E) = (B, 1, E)$ \\ & \hfill $\rightarrow (B, h \rightarrow H, S, A)$ & & \hfill $\rightarrow (B, h \rightarrow H, L, A)$ \\
Attention score (QK GEMM) & $(B, H, S, A) \times (B, H, A, S)$ & Attention score (QK GEMM) & $(B, H, 1, A) \times (B, H, A, L)$ \\ & \hfill $ = (B, H, S, S)$ & & \hfill $ = (B, H, 1, L)$ \\
Mask+Scale+SoftMax & $(B, H, S, S) \rightarrow (B, H, S, S)$ & Mask+Scale+SoftMax & $(B, H, 1, L) \rightarrow (B, H, 1, L)$ \\
Attention context (SV GEMM) & $(B, H, S, S) \times (B, H, S, A)$ & Attention context (SV GEMM) & $(B, H, 1, L) \times (B, H, L, A)$ \\ & \hfill $ = (B, H, S, A)$ & & \hfill $ = (B, H, 1, A)$ \\
Output projection (weight GEMM) & $(B, S, E) \times (E, E) = (B, S, E)$ & Output projection (weight GEMM) & $(B, 1, E) \times (E, E) = (B, 1, E)$ \\
Pre-normalization (RMSNorm) & Torch dimension is not changed & Pre-normalization (RMSNorm) & Torch dimension is not changed \\
Feed-forward & $w_2(\text{Act}(w_1(x)) \times w_3(x))$ & Feed-forward & $w_2(\text{Act}(w_1(x)) \times w_3(x))$ \\
Gate projection (weight GEMM) & $(B, S, E) \times (E, F) = (B, S, F)$ & Gate projection (weight GEMM) & $(B, 1, E) \times (E, F) = (B, 1, F)$ \\
Activation function & Torch dimension is not changed & Activation function & Torch dimension is not changed \\
Up projection (weight GEMM) & $(B, S, E) \times (E, F) = (B, S, F)$ & Up projection (weight GEMM) & $(B, 1, E) \times (E, F) = (B, 1, F)$ \\
Down projection (weight GEMM) & $(B, S, F) \times (F, E) = (B, S, E)$ & Down projection (weight GEMM) & $(B, 1, F) \times (F, E) = (B, 1, E)$ \\
Normalization (RMSNorm) & Torch dimension is not changed & Normalization (RMSNorm) & Torch dimension is not changed \\
Linear layer (weight GEMM) & $(B, S, E) \times (B, E, V) = (B, S, V)$ & Linear layer (weight GEMM) & $(B, 1, E) \times (B, E, V) = (B, 1, V)$\\
SoftMax & $(B, S, V) \rightarrow (B, S, V)$ & SoftMax & $(B, 1, V) \rightarrow (B, 1, V)$ \\
\midrule
\midrule
\textbf{Symbols} & \textbf{Description} & \textbf{Symbols} & \textbf{Description} \\
\midrule
\textbf{B} & Batch size & \textbf{S} & Sequence length \\
\textbf{V} & Vocabulary size & \textbf{E} & Embedding size \\
\textbf{H} & Head size of Query & \textbf{h} & Head size of Key and Value \\
\textbf{A} & E / H & \textbf{F} & Feed-forward dimension size\\
\textbf{C} & \# of generation stages & \textbf{L} & S + C = KV caching size \\
\bottomrule
\end{tabular}
\label{tab:sumgen layer}
\end{table*}

\section{Experimental Results}
In this section, two major experiments we conducted and the analysis details are described.
The experimental environment, the summarization and generation stage ratio, and the kernel analysis are addressed in Section \ref{Experiment Environment}, Section \ref{Summarization and Generation Stage Ratio}, and Section \ref{Kernel Analysis}, respectively.

\begin{table}[b]
\small
\caption{Experiment Environment}
\centering
\renewcommand{\arraystretch}{1.2}
\begin{tabular}{lllll}
\toprule
\textbf{Component} & \textbf{Specification} & & \\
\midrule
\textbf{GPU} & \textbf{NVIDIA RTX 6000 Ada 48GB} & Ada Lovelace Architecture & & \\
& NVIDIA CUDA Cores & 18,176 & & \\
& NVIDIA Tensor Cores & 568 & & \\
& NVIDIA RT Cores & 142 & & \\
& GPU memory & 48GB GDDR6 & & \\
& Memory bandwidth & 960 GB/s & & \\
& float32: 112 TFLOPS & bfloat16: 225 TFLOPS & RT Core: 210.6 TFLOPS & Tensor: 1457.0 TFLOPS \\
\midrule
\textbf{CPU} & AMD EPYC(TM) 75F3 CPU & @ 2.95GHz & & \\
\bottomrule
\end{tabular}
\label{tab:specs}
\end{table}

\subsection{Experiment Environment}\label{Experiment Environment}
We perform experiments using four LLM models, by sweeping input length, output length, and batch size.
As LLMs require highly computational and memory-intensive operations, we utilize a high-performance GPU and detailed environment settings are listed in Table \ref{tab:specs}.
To maintain a consistent experimental environment across four additional LLM configurations, we employ the ccdv/cnn\_dailymail dataset throughout the experiments \cite{nallapati2016abstractive}, which is provided by the TensorRT-LLM library.
In addition to analyzing the output results, we also perform GPU profiling to analyze the execution time and kernel configuration.

\begin{figure*}[!t]
  \centering
  \begin{subfigure}{0.24\textwidth}
    \centering
    \includegraphics[width=\columnwidth]{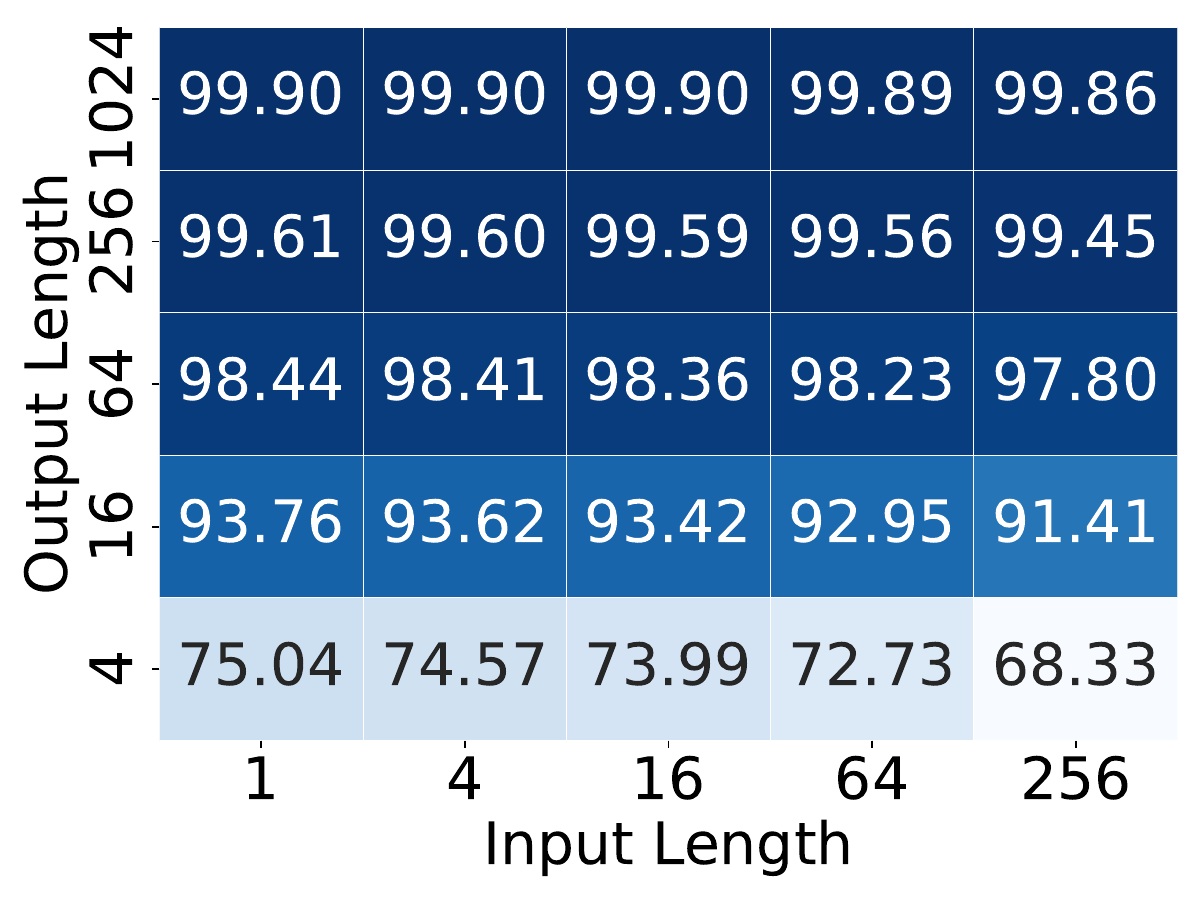}
    \caption{Llama2-7B (MHA), B = 1}
    \label{fig:image1}
  \end{subfigure}
  \hfill
  \begin{subfigure}{0.24\textwidth}
    \centering
    \includegraphics[width=\columnwidth]{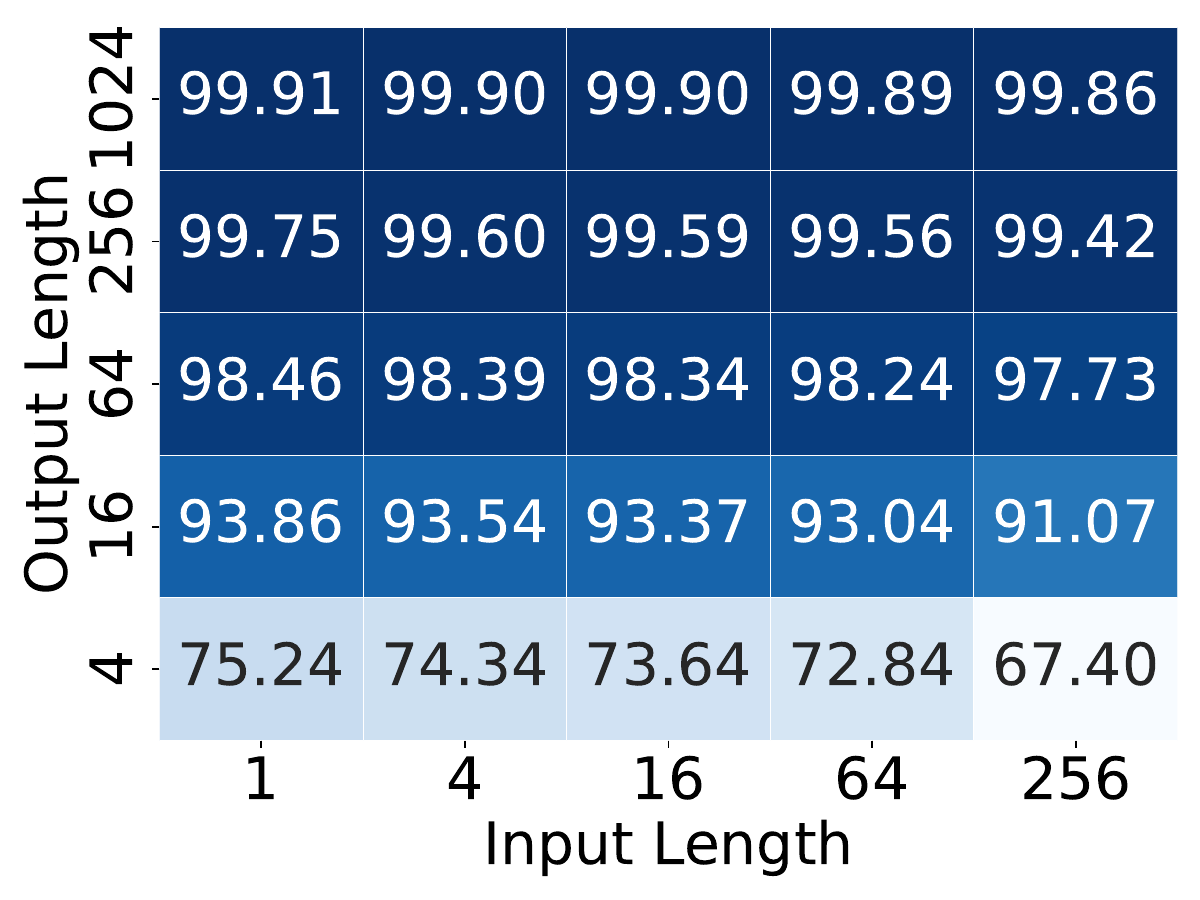}
    \caption{Gemma-7B (MHA), B = 1}
    \label{fig:image2}
  \end{subfigure}
  \hfill
  \begin{subfigure}{0.24\textwidth}
    \centering
    \includegraphics[width=\columnwidth]{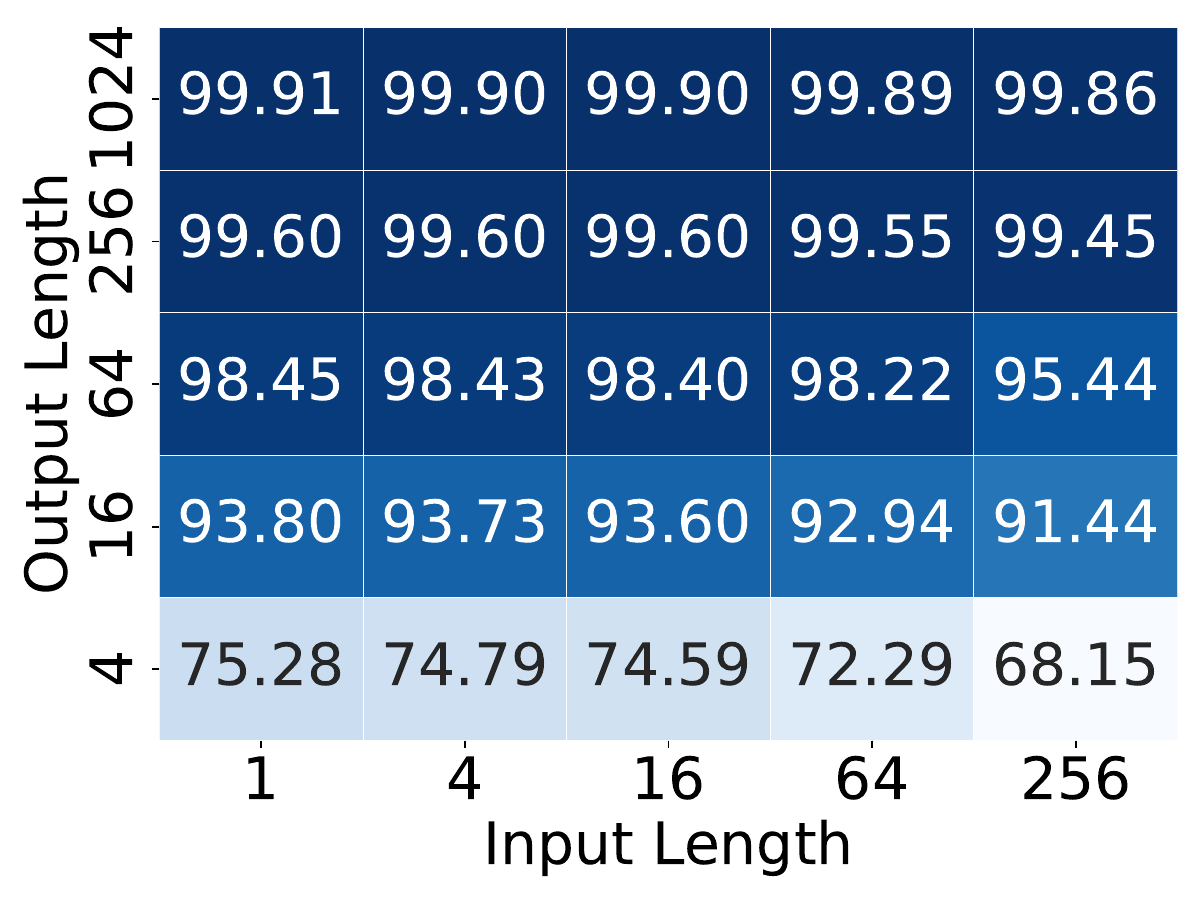}
    \caption{Llama3-8B (GQA), B = 1}
    \label{fig:image3}
  \end{subfigure}
  \hfill
  \begin{subfigure}{0.24\textwidth}
    \centering
    \includegraphics[width=\columnwidth]{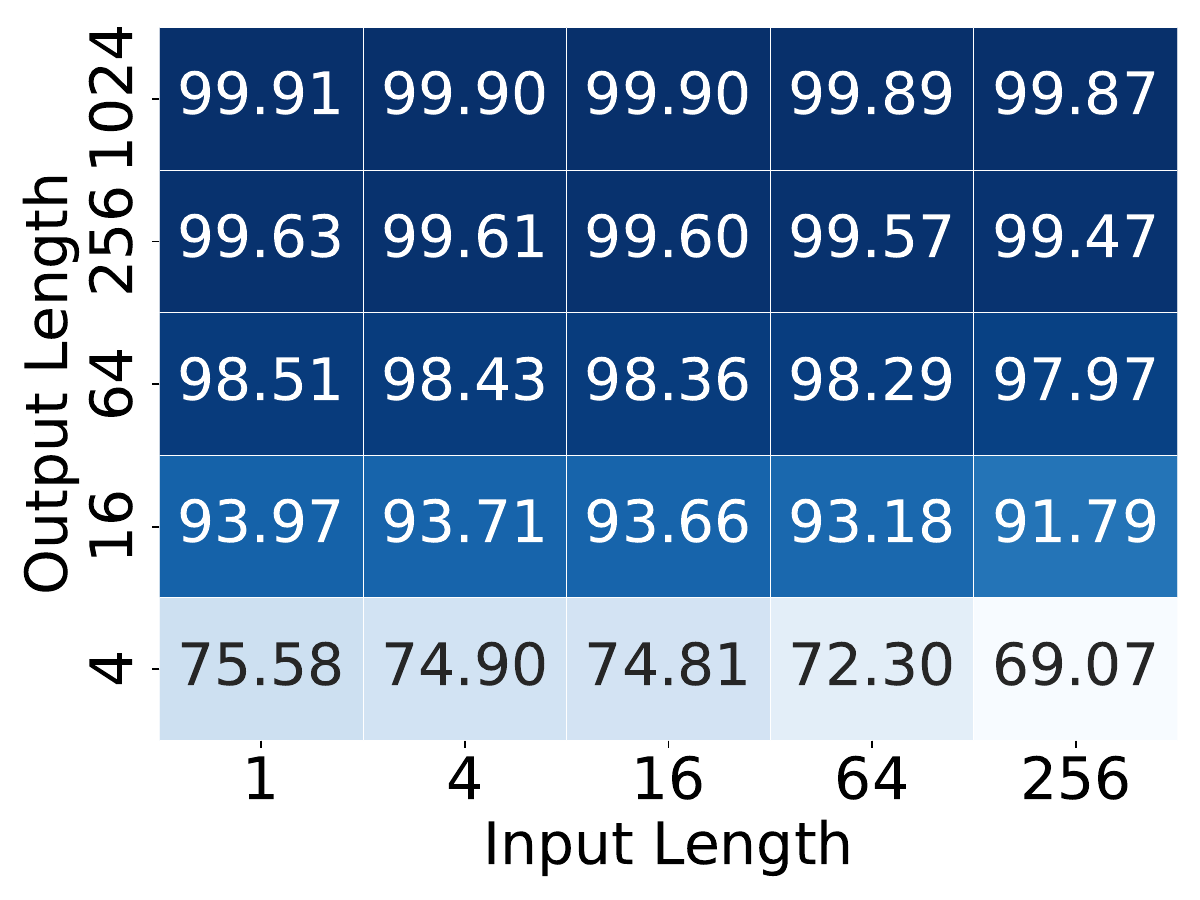}
    \caption{Gemma-2B (MQA), B = 1}
    \label{fig:image4}
  \end{subfigure}
  \\
  \vspace{10pt}
  \begin{subfigure}{0.24\textwidth}
    \centering
    \includegraphics[width=\columnwidth]{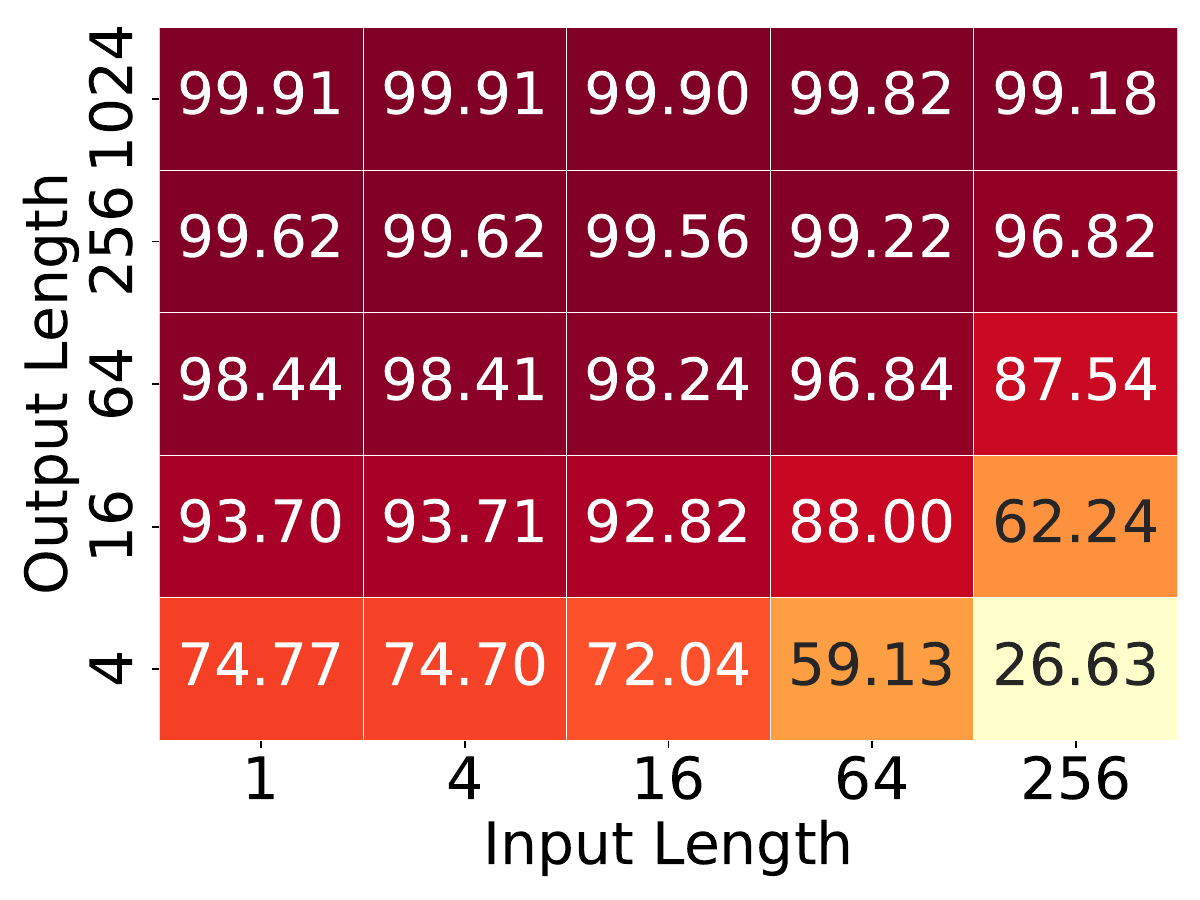}
    \caption{Llama2-7B (MHA), B = 8}
    \label{fig:image5}
  \end{subfigure}
  \hfill
  \begin{subfigure}{0.24\textwidth}
    \centering
    \includegraphics[width=\columnwidth]{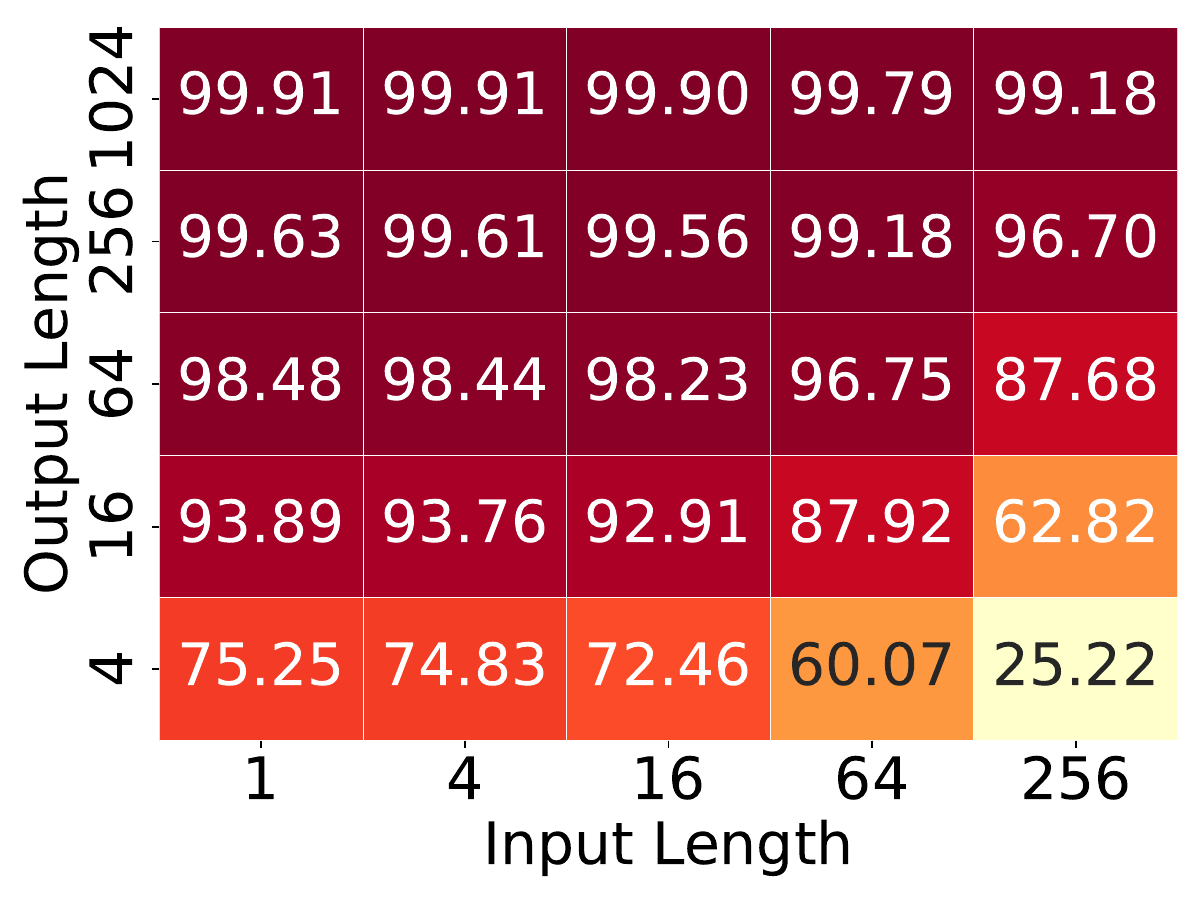}
    \caption{Gemma-7B (MHA), B = 8}
    \label{fig:image6}
  \end{subfigure}
  \hfill
  \begin{subfigure}{0.24\textwidth}
    \centering
    \includegraphics[width=\columnwidth]{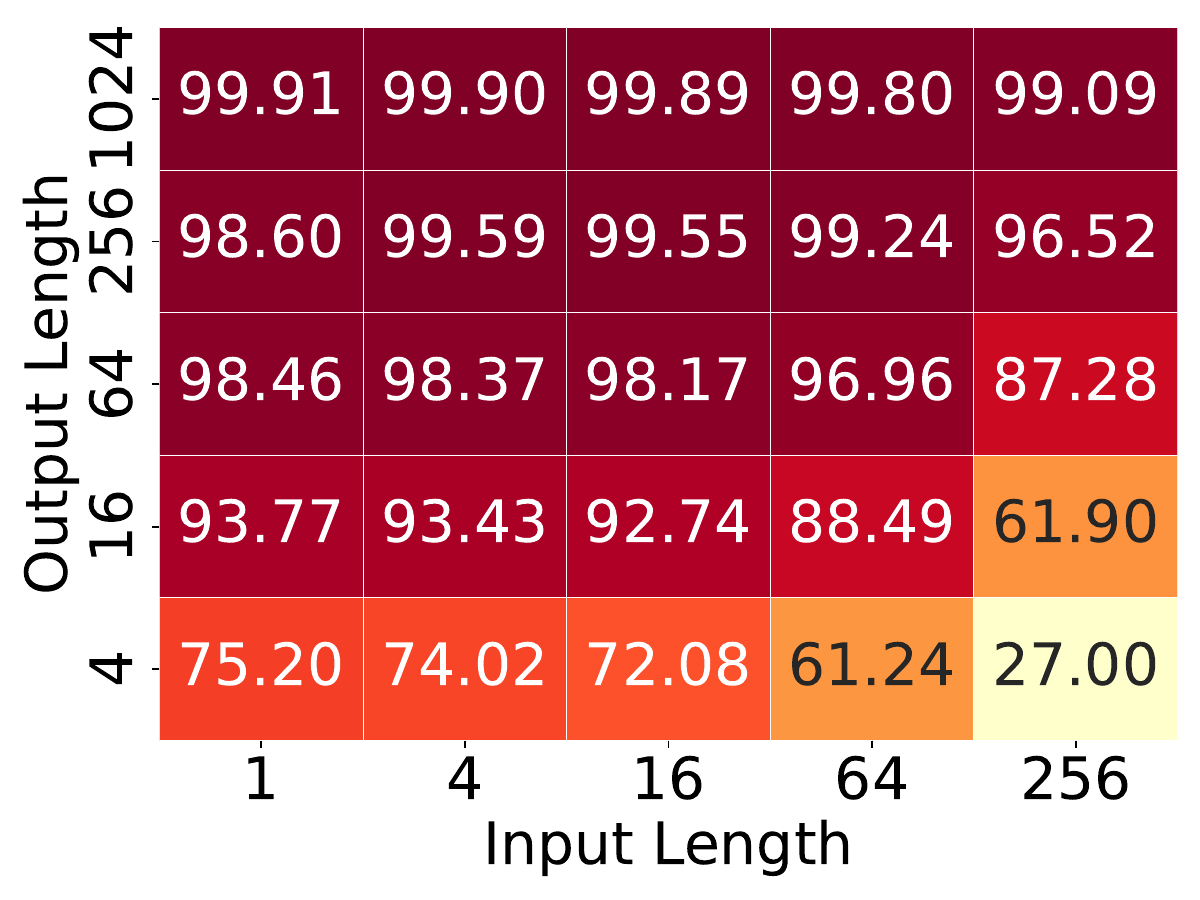}
    \caption{Llama3-8B (GQA), B = 1}
    \label{fig:image7}
  \end{subfigure}
  \hfill
  \begin{subfigure}{0.24\textwidth}
    \centering
    \includegraphics[width=\columnwidth]{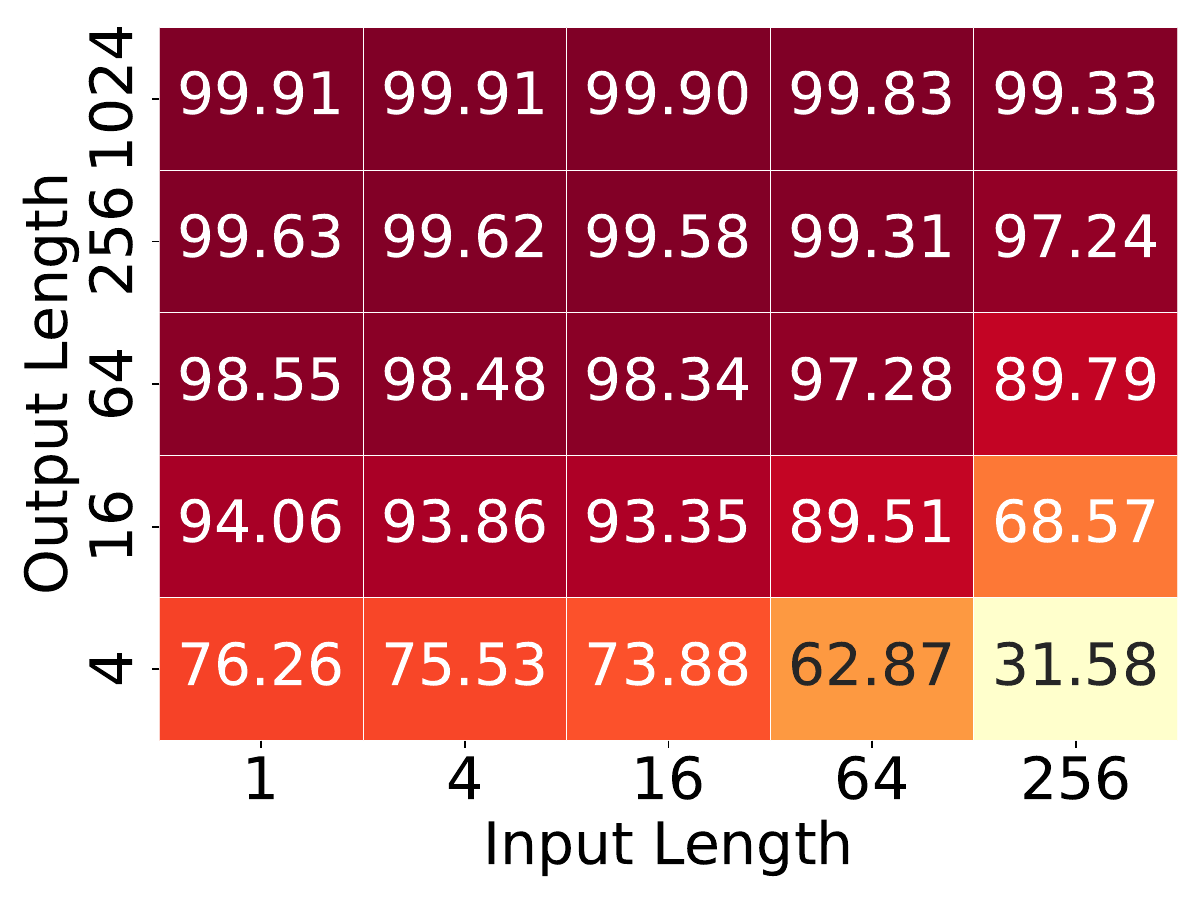}
    \caption{Gemma-2B (MQA), B = 1}
    \label{fig:image8}
  \end{subfigure}
  \caption{Proportion of Generation Stage of the total in case of Llama2-7B, Llama3-8B, Gemma-2B and Gemma-7B}
  \label{fig:inout}
\end{figure*}

\subsection{Summarization and Generation Stage Ratio}\label{Summarization and Generation Stage Ratio}
Figure \ref{fig:inout} illustrates the proportion of the generation stage execution time to the total execution time. 
It shows the results of sweeping the input lengths from 1 to 256 and the output lengths from 4 to 1024 with batch sizes of 1 and 8.
Furthermore, TensorRT-LLM is employed so that all models can be tested in the same environment as possible. 
The experiment is conducted on a basic TensorRT-LLM environment, including flash attention, and the data type is set to bfloat16. 
The heatmap analysis yields four major implications: Input sweep, output sweep, batch sweep and attention techniques, respectively.

{\bf{Input Sweep}}
As illustrated in all heatmaps in Figure \ref{fig:inout}, a consistent trend is observed where the proportion of the generation stage decreases as the input size increases.
This is because the arithmetic intensity of GEMM operations increases proportionally to the size of the sequence length (S) in the summarization stage. 
This trend becomes more evident as the length of the output sequence decreases, resulting in fewer instances of the generation stage.
 
{\bf{Output Sweep}} 
As the output length increases, the number of generation stage executions increases, resulting in an anticipated rise in the proportion of generation stage time relative to the total execution time.
This trend is observed throughout the results, where the proportion of the generation stage increases consistently with the length of the output.
Notably, when the input length is 1 and the output length is 4, the proportion of the generation stage is approximately 75\% in all cases. 
As shown in Table \ref{tab:sumgen layer}, which presents the dimensions calculated, the arithmetic intensity between the generation stage and the summarization stage remains nearly identical when the sequence length (S) is set to 1.
Therefore, the generation stage accounts for approximately three-fourths of the total runtime.
 
{\bf{Batch Sweep}}
However, the proportion of the generation stage reduces more rapidly with a larger batch size as the input size increases. 
This phenomenon arises because, with a large batch size, the GPU processes operations with an input size scaled by the batch size, leading to variations in sequence length and an increased number of operations for both summarization and generation stages.
In contrast to the generation stage, which transitions from GEMV to GEMM operations, the summarization stage intensifies arithmetic operations, making it more susceptible to the effects of increased input size.

Concerning internal operations, when the batch size is 1, the generation stage typically has operations per byte predominantly in the single-digit range. 
When the operations per byte are small, the operations tend to be memory-bound rather than compute-bound.
However, as the batch size grows, operations per byte gradually increase, leading to a shift towards compute-bound operations. 
Conversely, in the summarization stage, even with a batch size of 1, the operations per byte are significantly greater than 10, rendering all matrix multiplication operations nearly compute-bound.

{\bf{Attention Techniques}}
For models such as Llama2-7B and Gemma-7B, comparable results are observed due to the similar number of parameters. 
However, it appears that the generation stage ratio increases when transitioning from the two models using MHA to the Llama3-8B model using GQA and the Gemma-2B model using MQA. 
This is attributed to the relatively lower memory requirements for attention in models employing GQA and MQA, resulting in higher operations per byte compared to models employing MHA.
The observed reduction of memory usage in the attention blocks is identical at both summarization and generation stages and is approximately proportional to the change in the number of the KV heads.

The changes in operations per byte mentioned in the Batch Sweep are due to variations in torch size.
In contrast, the changes in operations per byte that occur with different types of attention blocks are due to variations in the amount of memory usage. 
From the perspective of high GPU utilization, both compute and memory must be optimized to avoid being bound to one side. 
As mentioned earlier in the batch sweep, the larger the operations per byte, the closer it is to the compute-bound, and the smaller it is, the closer it is to the memory bound.
From this perspective, it can be inferred that approaching the most optimized position of operations per byte could lead to a reduction in execution time.

However, what is illustrated in Figure \ref{fig:inout} represents the ratios rather than absolute values of the summarization and generation stages. 
Since the most optimized position varies across different models and batch sizes, identifying a consistent trend is challenging. 
Furthermore, the most optimized position is influenced by the performance of the computational units used.
Therefore, changes in the proportion of the summarization and generation stages resulting from different types of attention should be evaluated considering hardware, hyperparameters, parameters, and model architectures.

\subsection{Kernel Analysis}\label{Kernel Analysis}
\begin{figure*}[t]
\centering
\begin{subfigure}{0.49\textwidth}
  \centering
  \includegraphics[width=0.99\columnwidth]{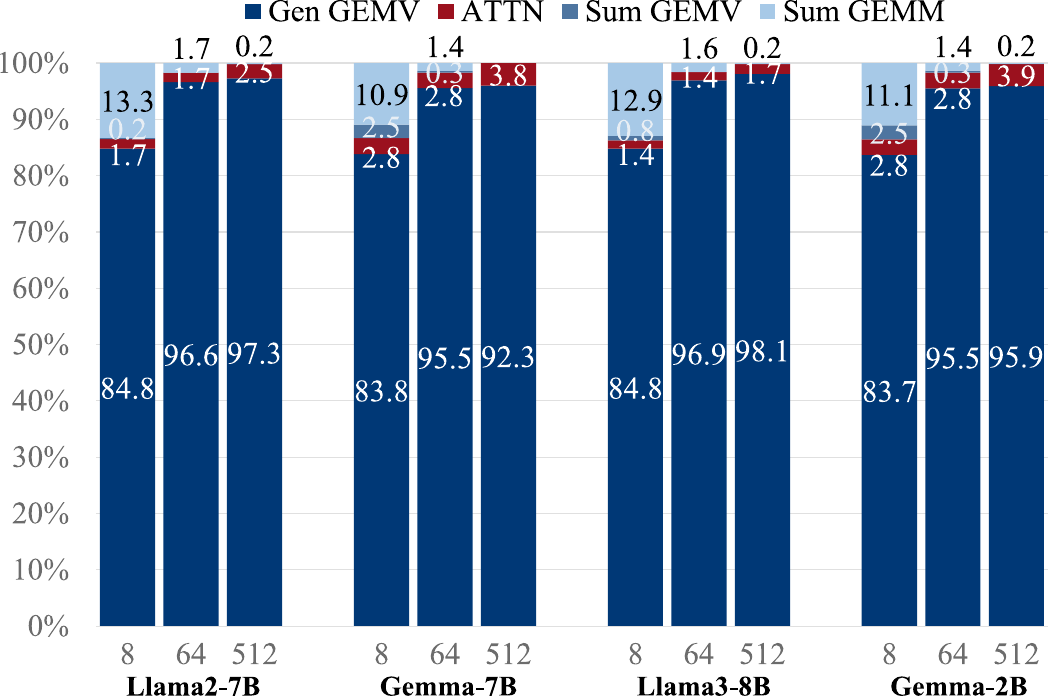}
  \caption{Kernel ratio, batch size = 1}
  \label{fig:batch1}
\end{subfigure}
\hfill
\begin{subfigure}{0.49\textwidth}
  \centering
  \includegraphics[width=0.99\columnwidth]{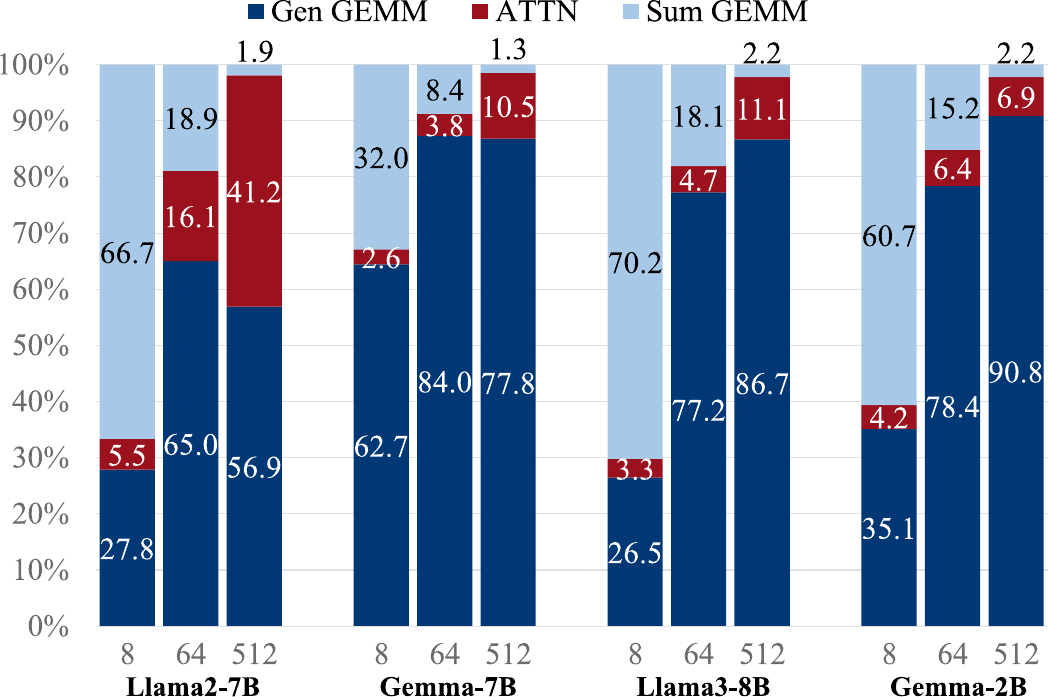}
  \caption{Kernel ratio, batch size = 64}
  \label{fig:batch64}
\end{subfigure}
\caption{Kernel Analysis of Llama2-7B, Gemma-7B (MHA), Llama3-8B (GQA) and Gemma-2B (MQA)}
\label{fig:batch sweep}
\end{figure*}

Figure \ref{fig:batch1} illustrates the proportion of each kernel in the overall execution time when the output length is 8, 64, and 512, while keeping the batch size at 1 and the input length at 64.
Figure \ref{fig:batch64} illustrates the proportion of kernels with the same output length variations (8, 64, and 512), but with a batch size of 64 and an input length of 64.
Both figures analyze four different models within the same kernel categories, as listed below. 
The analyzed kernels account for more than 95\% of the total execution time and they are categorized into five groups: Generation stage GEneral Matrix-Vector multiplication (Gen GEMV), Summarization stage GEMV (Sum GEMV), Attention blocks (ATTN), Generation stage GEneral Matrix-Matrix multiplication (Gen GEMM), and Summarization stage GEMM (Sum GEMM).

{\bf{Kernel Characteristics}}
When the batch size is 1, the proportion of summarization and generation stages based on input and output sweeps reveals that GEMM operations dominate the summarization stage, while GEMV operations dominate the generation stage across all four models.
However, when the batch size is 64, the input sequences are multiplied by 64, resulting in GEMM operations becoming the primary kernels for both the summarization and generation stages across all models.
The execution times of the attention mechanism in both the generation and summarization stages are combined and represented as the ATTN component, as shown in Figure \ref{fig:batch sweep}.

{\bf{Single Batch}}
Regardless of the batch size, the graph demonstrates that the execution time of the generation stage increases with the output length.
This graph illustrates the kernel composition and its impact on changes in execution time.
The experiments are conducted with two different batch sizes.
In edge environments, where it is common to process a single execution at a time, the scenario with a batch size of 1 can be considered representative.

In operations with a batch size of 1, GEMV operations account for more than 80\% of the total execution time for all models and even exceed 95\% when the output is over 64.
Consequently, a hardware configuration optimized for GEMV operations is essential when the output length is greater than 64.
The GPU utilization of the GEMV kernels reveals that the memory bandwidth usage is relatively low compared to the computational capacity of the kernels, indicating that the operations are memory-bound.

{\bf{Large Batch}}
In environments such as servers, where multiple inputs are processed simultaneously, a larger batch size can be utilized to handle inputs concurrently.
When the batch size is 64, the calculation is performed by multiplying the sequence length by the batch size, resulting in the GEMM kernels occupying the dominant part of the execution time. 
In contrast to the case where the batch size is 1, it can be observed that the proportion of attention in the total execution time increases significantly. 
As the size of the KV increases, the generation stage becomes more computationally intensive, leading to a greater demand for both computation and memory resources.
This is evidenced by the observation that as the batch size increases from 1 to 64, the amount of computation required to generate an output token increases.

{\bf{Attention Techniques}}
The rate of increase varies depending on which attention mechanism is employed. 
As previously stated in \ref{Summarization and Generation Stage Ratio}, the low-memory attention mechanism exhibits a relatively high operation per byte ratio in attention blocks.
Consequently, it can be observed that in both Llama and Gemma models, the trend of increasing the ratio of attention execution time in GQA and MQA is relatively modest.

\section{Conclusion}
The rationale for employing TensorRT-LLM in this analysis is that it leverages a plethora of existing acceleration techniques, enabling consistent acceleration and analysis of the model.
In particular, on the issue of attention, the difference in kernel execution time is significant before and after the application of kernel fusion and attention acceleration techniques.
Before this, attention blocks were not well suited to GPUs, resulting in occupying an excessively large portion of the total execution time.
Therefore, there is considerable discussion on how to accelerate attention. In this analysis, we employ flash attention, a feature included in the TensorRT-LLM backend, to achieve the desired outcome in a consistent environment.

The preceding analysis has revealed which models exhibit the desired behavior in specific hyperparameter settings and how they function in such contexts.
It is important to note that the types of kernels used are completely different in small batch size of edge environments compared to large batch size of server environments. 
Furthermore, the predominant kernel also varies significantly depending on the required output length.

In the context of edge AI, the GEMV operation represents a significant proportion of the total execution time. 
In contrast, for server-based LLM environments, the most significant challenge has been reducing the execution time of the Attention mechanism.
To address this, various attention architectures and mechanisms are employed to program and reduce the amount of memory derived from an attention part. 
Our evaluations indicate that by employing acceleration techniques and advanced attention methods including GQA and MQA, we can mitigate the computational burden of attention to a certain extent.

Consequently, our survey and evaluation demonstrate that different types of hardware, distinct from the previously utilized GPU, are required depending on the specified hyperparameters.

\section{Acknowledgements}
We thank the DATES Lab members for their support and feedback during this work. This work is supported by Backdrop Build Lab and RunPod Cloud Service.

\vfill

\end{document}